# LaDA: Latent Dialogue Action For Zero-shot Cross-lingual Neural Network Language Modeling

Zhanyu Ma (mazhanyu21s@ict.ac.cn)

Shuang Cheng (chengs18@tsinghua.org.cn)

Jian Ye [1] (jye@ict.ac.cn)
Institute of Computing Technology, Chinese Academy of Sciences
University of Chinese Academy of Sciences
Beijing Key Laboratory of Mobile Computing and Pervasive Device
Beijing, China

## Abstract

Cross-lingual adaptation has shown effectiveness in low-resource spoken language understanding (SLU) systems. However, existing approaches often show unsatisfactory performance on intent detection and slot filling, especially for distant languages, which are dramatically different from source language in scripts, morphology, or syntax. To solve the aforementioned challenges, we propose Latent Dialogue Action (LaDA) layer to optimize decoding strategy. The model consists of an extra latent dialogue action layer. It enables our model to enhance a system's ability to handle conversations in complex multi-lingual intent and slot value of distant languages. To our best knowledge, our work is the first comprehensive study of the use of latent variables for cross-lingual SLU policy optimization during decode stage. LaDA achieves state-of-the-art results in zero-shot and few-shot adaptation both on public datasets.

**Keywords:** Artificial Intelligence; Computer Science; Language understanding; Machine learning; Cross-linguistic analysis; Neural Networks

## Introduction

In spoken language understanding (SLU) systems, data-driven neural-based supervised training strategies have proven successful (Shon et al., 2022; Lau, Fyshe, & Waxman, 2021; Misra, Ettinger, & Rayz, 2021). However, gathering large quantities of high-quality training data is not only costly but also time-consuming, making these techniques inapplicable to low-resource languages owing to a lack of training data (Gritta, Hu, & Iacobacci, 2022; Wu, Ding, Yang, & Li, 2022). Cross-lingual adaptation has evolved organically to address this challenge (Matzen, Ting, & Stites, 2021; Frank, 2021), using training data in high-resource source languages while minimizing the need for training examples in low-resource target languages.

In general, the first challenge in cross-lingual adaptation is that imperfect alignment of word-level and sentence-level representations between the source and target language limits the adaptation performance (Frank, 2021). Existing cross-lingual transfer learning approaches rely mostly on pre-trained cross-lingual word embeddings or contextual models, which represent natural language texts with comparable meanings in different languages in close proximity to each other in a common vector space (Lu, Huang, Qu, Wei, & Ma, 2022).

Unfortunately, those approaches often show good performance on intent detection in a few languages (Louvan & Magnini, 2020). The results on slot filling and intent detection, however, are not satisfactory, especially for some languages at great distances, which are dramatically different from English in scripts, morphology, or syntaxes, such as Hindi (hi) and Thai (th) (Tjandra et al., 2022). Similarly, even though we assume that the alignment is perfect, the trained model generations still suffer from intent/slot conflict, repetitiveness and contradictions owing to grammatical and syntactical variances across languages. Obviously, focusing on alignment of representations learning alone is not enough.

Therefore, we further argue the second challenge that current cross-lingual models do not adequately understand the deeper meaning of slots generations and frequently contradict their own intents. Only focusing on multi-language representation learning will lose the effect of model generation and semantic parsing is also significant in the decoding stage.

For conventional modular systems, the slot probabilistic space is defined by hand-crafted semantic representations such as slot-values, and the objective is to establish a dialog policy that selects the optimal hand-crafted values at each dialog turn (Chai, Liang, & Duan, 2022). But it is limited because it can only handle simple multi-lingual intent domains whose entire slot space can be captured by hand-crafted representations in similar language scripts, morphology, or syntax (Zhao, Xie, & Eskenazi, 2019; Arora, Shuster, Sukhbaatar, & Weston, 2022). Meanwhile, this cripples a system's ability to handle conversations in complex multi-lingual intent and slot value, especially when significant differences in grammatical and syntactical variances across languages. Some recent works show that slot and intent are highly correlated and have similar semantic meanings in a sentence (Ma, Ye, Yang, & Liu, 2022; Qin, Li, Che, Ni, & Liu, 2021). Intent recognition can help with slot filling, they propose joint models to consider the correlation between these two tasks. However, the error of intent recognition will continue until the slot information is filled. Even if the intent is correctly identified, the slot information has a high probability of be-ing wrong.

---
[1] Corresponding Author

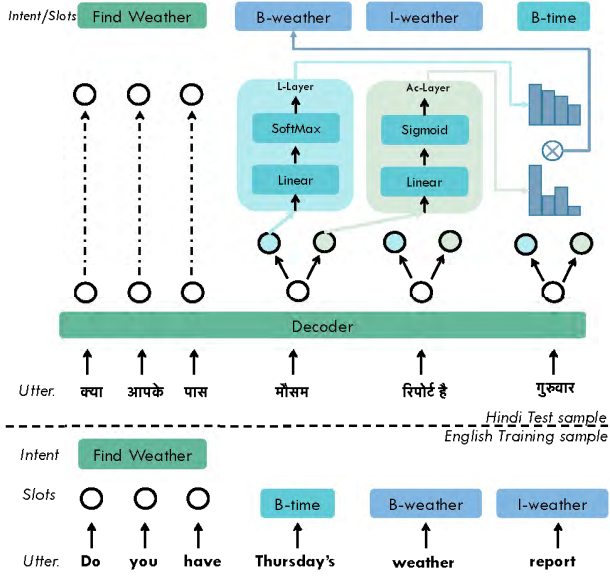

Figure 1: Cross-lingual SLU task and the framework of LaDA. Ac-Layer donates latent dialogue action and L-Layer donates language modeling output.

In this paper, we are concerned with the problem of conflict between intent and slot information in cross-linguistic contexts where there are significant grammatical differences and improve the performance of SLU task both in few-shot and zero-shot scenarios. To solve the aforementioned challenges, we present a new model architecture named Latent Dialogue Action (LaDA) that is capable of training on supervised data indicating slot filling. Prior work has shown that learning with latent variables leads to benefits like multilingual representation (Bao, He, Wang, Wu, & Wang, 2020; Liu, Winata, Xu, Lin, & Fung, 2020; Liu et al., 2019; Ma, Liu, & Ye, 2023).

The model consists of a standard decoder architecture with an extra latent dialogue action layer for each output token, in addition to the usual language modeling output, as shown Fig. 1. Both the language modeling layer and the latent dialogue action layer are trained using labeled data, with the bulk of decoder parameters being shared across the two tasks. To our best knowledge, our work is the first comprehensive study of the use of latent variables for slot filling policy optimization in cross-lingual spoken language understanding.

Experiments demonstrate our model enhances a system's ability to handle conversations in complex multi-lingual intent and slot value, especially when significant differences in grammatical and syntactical variances across languages, improving adaption robustness. And LaDA achieves state-of-the-art results in zero-shot and few-shot adaptation of German, Spanish, French, Hindi and Thai for the natural spoken language understanding task (i.e., the intent detection and slot filling).

## Model

In this section, we will introduce the LaDA model. We will start by laying out the notation and background of language modeling and then introduce our new architecture.

### Problem Formulation

NLU is primarily responsible for intent detection and slot filling. Given $L$ words $u = [w_1, w_2, \ldots, w_L]$ as well as a collection of predetermined intent types $I$ and slots types $S$, The purpose of the intent detection is to attempt to anticipate the intent. $o^I \in I$ based on the utterance $u$ while the slot filling is a sequence tagging problem, mapping the word sequence $[w_1, w_2, \ldots, w_L]$ into semantic slots $[s_1, s_2, \ldots, s_L]$ where $s \in S$.

Few-shot learning (FSL) extracts prior experience that allows quick adaption to new problems. Therefore, FSL models are usually first trained on a set of source language, then evaluated on another set of unseen target language.

A target language only contains a few labeled examples, which is called support set $\mathcal{S} = \left\{(x^{(i)}, y^{(i)})\right\}_{i=1}^{|\mathcal{S}|}$. $\mathcal{S}$ includes $K$ examples (K-shot) for each of $N$ classes (N-way). Taking classification problem as an instance: given an input query example $x = \langle x_1, x_2, \ldots, x_n \rangle$ and a K-shot support set $\mathcal{S}$ as references, we find the most appropriate class $y^*$ of $x$:

$$y^* = \arg\max_y \ p(y \mid x, \mathcal{S}). \tag{1}$$

Actually, zero-shot learning is a limit of few-shot learning, which further tests the cross-lingual ability of the model. In zero-shot setting, support set $\mathcal{S}$ is NULL, directly use the trained cross-lingual model to infer the target language that is not in the training set.

### Intent detection and slot filling

Intent detection and slot filling are two key tasks of SLU. Given the the word sequence $u = w_1, w_2, \ldots, w_L$ are passed into multi-lingual language model [2] [3] [4] and obtain derive sentence representations $h^u$ and contextual embeddings $e_1, e_2, \ldots, e_{L-1}, e_L$. To jointly learn both tasks, the objective function $\mathcal{L}$ is formulated as:

$$\mathcal{L}_{SLU} = -\sum_{i=1}^{n^I} \hat{y}_i^I log(y_i^I) - \sum_{j=1}^{L}\sum_{i=1}^{n^S} \hat{y}_{j,i}^S log(y_{j,i}^S) \tag{2}$$

$n^I$ is the number of intent types and $\hat{y}_i^I$ is the gold intent label. $L$ is the sequence length of a sentence, $n^S$ means the number of slot types as well as $\hat{y}_j^S$ is the gold slot label.

$$y^I = Softmax(W_h^I h^u + b_h^I) \tag{3}$$

$y^I$ is the intent prediction distribution and $W_h^I, b_h^I$ are trainable parameters. The slot distribution for each word can be predicted as $y_j^S$

$$y_j^S = Softmax(W_h^S h_j^S + b_h^S), j \in \{1, \ldots, L\} \tag{4}$$

---
[2] https://huggingface.co/bert-base-multilingual-cased
[3] https://github.com/facebookresearch/XLM/
[4] https://github.com/facebookresearch/LASER

and $W_h^S, b_h^S$ are trainable parameters. Here $h_j^S$ is a combined representation with a word embedding $e_j$ and the intent embedding $h^I$,

$$h_j^S = \text{Average}(e_j, h^I). \tag{5}$$

## LaDA Language Model

Let word sequence of $U$, $u = w_1, w_2, ..., w_L$ be supervised training data where each token sequence $x_1, ..., x_T$ is labeled. This is accomplished either by marking the whole sequence with a class $y = c$ or, In the finely grained scenario, each token has a class label, giving $y_1, ..., y_T$. Then the goal is to learn to generate conditioned on a specified given class, which means modeling $P(x_t|x_1, ..., x_{t-1}, y_t)$. Using conditional generation probability, the objective function $P_{LaDA}$ is:

$$P(x_t|x_1, ..., x_{t-1}, y_t) \propto P(x_t|x_1, ..., x_{t-1})P(y_t|x_1, ..., x_t). \tag{6}$$

The first term could be calculated using a language model, but the second needs a classifier that minimizes the cross-entropy loss:

$$\mathcal{L}_{\text{action}} = -\log P(y_t = c|x_1, ..., x_t). \tag{7}$$

We thus propose LaDA, a model that combines conventional language modeling with a latent dialogue action layer. This enables efficient training of the model using supervised data. Then, at the moment of inference, we can produce based on required properties (positive class labels).

As shown in Fig. 1, Initial processing of input tokens is performed by a shared autoregressive core, for which a transformer decoder was used in our studies. The processed token representations are then delivered to two separate levels. The first layer is a conventional language modeling layer consisting of a linear layer followed by a softmax to generate a multinomial distribution. This layer is trained by optimizing loss $L_{SLU}$.

The second layer is for dialogue action and it also maps each token representation into a $|y_j^S|$ dimensional vector using a linear layer. Then, however, a *sigmoid* is applied to each word to generate an independent binomial distribution.

$$y_j^S = Sigmoid(W_{ac}^S h_j^S + b_{ac}^S) = \frac{exp(W_{ac}^S h_j^S + b_{ac}^S)}{exp(W_{ac}^S h_j^S + b_{ac}^S) + 1} \tag{8}$$

Where $W_{ac}^S$ and $b_{ac}^S$ are trainable parameters. Note that while tokens $x_1, ..., x_{t-1}$ are inputted and processed by the shared transformer core, the next token candidates for $x_t$ are encoded in the row vectors of the linear layer in the latent dialogue action layer. This layer optimizes loss $\mathcal{L}_{\text{action}}$ from Eq. 7.

The final objective function is formulated as:

$$\mathcal{L} = \mathcal{L}_{SLU} + \alpha \mathcal{L}_{\text{action}}, \tag{9}$$

where $\alpha$ is a hyperparameter weighting the dialogue action loss.

To generate a sequence conditioned on a certain class $c$ according to Eq. 6, we aggregate the outputs from the two layers to determine the probability of the next token

$$P(x_t) = P(x_t) P_{\text{action}}^\alpha (y_t = c), \tag{10}$$

Adjusting the parameter $\alpha$ at the time of inference allows us to change the weight of the latent dialogue action relative to the language model layer, where $\alpha = 0$ reverts to traditional language modeling. Tokens are generated in the same way as traditional language models, from left to right.

Two characteristics make LaDA's unified design efficient: 1. Because the latent dialogue action is autoregressive rather than bidirectional, the computations of previous token representations may be used for future token classifications rather than processing the complete sequence $x_1, ..., x_t$ at each time step $t$. 2. Instead of categorizing each token candidate $x_t$ individually, the latent dialogue action layer classifies them all simultaneously. In large transformers, executing it once requires the same processing power as the typical language modeling layer. LaDA's computing efficiency during training and inference is comparable to running the language model alone.

---

**Algorithm 1** The LaDA algorithm.
---
**Input:** Training data tuples in English; inference data tuples in target language;
**Output:** Intent and slot value in English; Intent and slot value in target language as shown in **figure 1**.
1: $PXLM \leftarrow$ Pretrained Cross-lingual LM
2: $X \leftarrow$ Training data tuples in English
3: # training loop
4: **for** $(x_t, y) \in X$ **do**
5: $\quad E_t \leftarrow PXLM(x_t)$ #get semantic embedding
6: $\quad P(x_t) \leftarrow L - Layer(E_t)$ #refer Eq. 1 and Fig. 1
7: $\quad P_{\text{action}}^\alpha \leftarrow Ac - Layer(E_t)$ #refer Eq. 8 and Fig. 1
8: $\quad \hat{y} \leftarrow P(x_t) P_{\text{action}}^\alpha$ #refer Eq. 10
9: $\quad total\_loss \leftarrow task\_loss\_fn(\hat{y}, y)$ #refer Eq. 9
10: $\quad$ # update model parameters
11: **end for**

---

## Experiments

### Implementable Details

We introduce several competitive baselines, including CoSDA-ML (Qin, Ni, Zhang, & Che, 2021), ORT (Liu et al., 2021), SLU-LR (Liu et al., 2020), BiLSTEM with CRF (LVM) (Liu et al., 2019), XLM (Conneau & Lample, 2019) and mBERT (Pires, Schlinger, & Garrette, 2019). we adopt classification accuracy (Acc.) to evaluate the performance of intent detection while using F1 score to measure the performance of slot fillings.

Our investigations employ WordPiece embeddings with a 110k-token vocabulary (Devlin, Chang, Lee, & Toutanova, 2019). LASER produces 1024-dimensional sentence embeddings, whereas mBERT generates 768-dimensional word embeddings.

Notice that we take the first subword embedding as word-level representation following (Qin, Ni, et al., 2021) while incorporating mBERT for slot filling. We also set the size of intent embedding to 768, then train our model for 9 epochs with a batch size of 64 and a learning rate of 0.002. α in Eq. 9 we set it as 0.125. We adopt AdamW (Loshchilov & Hutter, 2018) to optimize our LaDA and select the hyper-parameters by beam search (*nums_beam* = 4). To be noticed, we choose the best model reported by validation split from English dialog data. Besides, we adopt the gold intent instead of the predicted intent $o^I$ in equation: $o^I = Argmax(y^I)$ to guide the slot filling during the early stages of training period to avoid error continuation. We fine-tune mBERT using the sequence labeling task and then apply it to the slot filling job following (Pires et al., 2019)

Table 1: Acc comparison on MTOP between different cross-lingual approaches. We highlight the best scores in bold and underline the second best for each language.

| Model | Test on MTOP: Intent Acc | | | | |
|---|---|---|---|---|---|
| | de | es | fr | hi | th |
| Zero-shot Learning | | | | | |
| ORT | 72.91 | 69.34 | 66.54 | 43.18 | 44.05 |
| SLU+LR&ALVM | 72.73 | 71.09 | 66.22 | 41.69 | 46.77 |
| BiLSTM with CRF | 72.43 | 69.90 | 66.88 | 46.70 | 46.19 |
| BiLSTM with LVM | 68.52 | 68.00 | 62.62 | 37.10 | 52.36 |
| mBERT | 68.29 | 65.02 | 66.18 | 22.76 | 24.77 |
| XLM | 58.61 | 58.46 | 65.24 | 36.65 | 34.22 |
| mBERT+CoSDA-ML | 88.70 | 84.45 | 85.95 | 63.23 | 68.81 |
| XLM+CoSDA-ML | 77.13 | 76.93 | 85.85 | 65.46 | 61.12 |
| LaDA | **90.44** | **90.87** | **89.48** | **75.20** | **77.48** |
| w/o LaDA | 81.43 | 80.18 | 82.63 | 65.24 | 68.00 |
| w/o LASER, w/ mBERT | 82.07 | 83.61 | 80.59 | 69.32 | 69.26 |
| w/o LASER, w/ XLM | 81.41 | 81.09 | 82.30 | 71.00 | 73.41 |
| w/o mBERT, w/ XLM | 87.22 | 88.36 | 86.44 | 73.61 | 75.30 |

Table 2: F1 comparison on MTOP between different cross-lingual approaches. We highlight the best scores in bold and underline the second best for each language.

| Model | Test on MTOP: Slot F1 | | | | |
|---|---|---|---|---|---|
| | de | es | fr | hi | th |
| Zero-shot Learning | | | | | |
| ORT | 51.42 | 36.17 | 32.75 | 20.63 | 18.31 |
| SLU+LR&ALVM | 52.30 | 37.88 | 34.17 | 21.28 | 19.67 |
| BiLSTM with CRF | 43.19 | 29.12 | 30.27 | 20.13 | 16.71 |
| BiLSTM with LVM | 45.73 | 34.05 | 32.07 | 17.12 | 19.62 |
| mBERT | 45.25 | 42.88 | 43.99 | 13.43 | 9.38 |
| XLM | 36.99 | 23.23 | 34.76 | 15.99 | 5.28 |
| mBERT+CoSDA-ML | 74.19 | 70.30 | 72.12 | 47.99 | 33.50 |
| XLM+CoSDA-ML | 64.90 | 40.77 | 61.00 | 42.10 | 13.91 |
| LaDA | **75.12** | **73.92** | **74.63** | **50.60** | **37.46** |
| w/o LaDA | 67.42 | 63.90 | 70.56 | 45.81 | 31.40 |
| w/o LASER, w/ mBERT | 70.21 | 66.21 | 69.80 | 45.29 | 33.62 |
| w/o LASER, w/ XLM | 69.42 | 64.93 | 67.30 | 44.00 | 32.74 |
| w/o mBERT, w/ XLM | 72.61 | 70.88 | 74.02 | 48.45 | 35.14 |

### Datasets

**MTOP** is a Multilingual Task-Oriented Parsing dataset provided by (Li et al., 2021) that covers interactions with a personal assistant (intent recognition and slot filling tasks). We use the standard flat version. A compositional version of the data designed for nested queries is also provided. MTOP contains 100K+ human-translated examples in 6 languages (en, de, es, fr, th, hi) and 11 domains.

### Baselines

We introduce several competitive baselines, including CoSDA-ML (Qin, Ni, et al., 2021), SLU-LR (Liu et al., 2020), BiLSTEM with CRF (LVM) (Liu et al., 2019), XLM (Conneau & Lample, 2019) and mBERT (Pires et al., 2019).

### Metrics

The different performance evaluation metrics (PEM) used for quantifying our method are accuracy (Acc.), precision (P), recall (R), F1 score (F1), and their micro-average weighted values. The LaDA's predictions can be divided into TP (true-positive), TN (true-negative), FP (false-positive), and FN (false-negative). Here, TP/FP (and TN/FN) signify the cardinality of correct/incorrect predictions for the positive (and negative) class respectively. Based on TP, TN, FP, and FN values, the PEMs are defined as: Acc. $= \frac{TP+TP}{TP+TN+FP+FN}$, $P = \frac{TP}{TP+FP}, R = \frac{TN}{TN+FN}$, and $F1 = \frac{2TP}{2TP+FP+FN}$.

### Zero-shot setting

Table 1, 2 show that LaDA improved the state-of-the-art performance in the zero-shot environment, and our zero-shot best German, Spanish, French, Hindi and Thai model's intent detection and slot filling performance is comparable to the strong baseline CoSDA-ML 1%-few-shot model, which uses some multilingual resources. LaDA enhances a system's ability to handle conversations in complex multi-lingual intent and slot value, especially when significant differences in grammatical and syntactical variances across languages, improving adaption robustness. Self-supervised training with the LaDA also disentangles latent variables for various slot types, increasing robustness. Even though Thai intention recognition we have improved by more than 50% from mBERT, but Thai slot filling performance is limited. We hypothesize there is not enough Thai corpus in the cross-lingual pre-trained model data.

To sum up, On the zero-shot scenario for MTOP, our model outperforms CoSDA-ML which is the former SOTA model in intent accuracy/f1-score by 1.74%/0.93% in German, 6.42%/3.62% in Spanish, 3.53%/2.51% in French, 9.74%/2.61% in Hindi, 8.67%/3.69% in Thai.

### Ablation Study

To investigate the effects of individual component, we conduct ablation study and report the results in Table 1, 2.

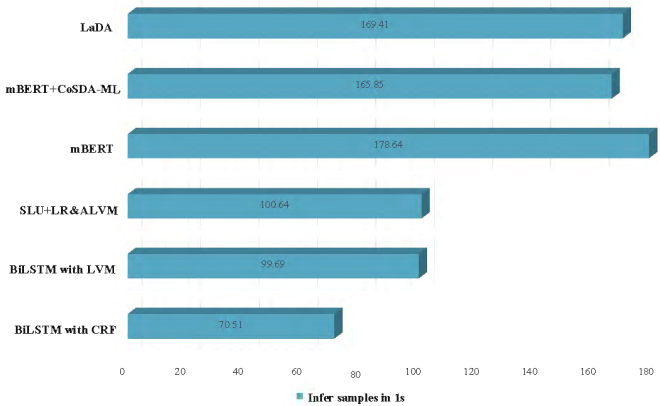

Figure 2: Inference speed of LaDA vs. baselines.

Firstly, we remove the latent dialogue action from the proposed model and find that it would perform worse without the guidance from latent action. It drops 9.01/7.70 in German, 10.69/10.02 in Spanish, 6.85/4.07 in French, 9.96/4.79 in Hindi, 9.48/6.06 in Thai, for intent detection /slot filling respectively in MTOP. It is probably because latent action can provide related information and help to find more accurate semantic slots.

Secondly, we investigate the importance of sentence alignments by replacing LASER with two pre-trained language models, i.e., mBERT or XLM, to derive the sentence embeddings. It validates the effectiveness of LASER that the performance of models with either mBERT or XLM is inferior to that with LASER which can produce aligned sentence representations in different languages. In addition, we also find that mBERT can produce better token-level embeddings for slot filling than XLM, thus we derive word-embedding with mBERT in LaDA.

## Inference Speed

The inference speed of various models is shown in Figure 2. The experimental environment is Intel(R) Xeon(R) Silver 4214R CPU@2.40GHz and NVIDIA Quadro GP100(16G). LaDA has only one additional latent dialogue action layer per token, but otherwise it is the same size as the baseline and thus generates almost the same number of samples per second. Even though separate models must be used for encoding and classification, the reorderer operating on beam search candidates does not cause much slowdown.

## Conclusion

In this paper, we propose a new network structure: latent dialogue action layer, which acts in the decoding stage of conditional generative models. Extensive experimental results show that our model enhances a system's ability to improve adaption robustness and handle conversations in complex multi-lingual intent and slot value, especially when significant differences in grammatical and syntactical variances across languages. Our proposed method is independent of the backbone network (e.g. mBERT, XLM model) and is highly adaptable. Actually, reinforcement learning can also be used as an action in the intent and slot state space, and we will continue to explore this aspect. As future work, we plan to investigate the performance of our method on different cross-lingual tasks.

## Acknowledgements

The research work is supported by National Key RD Program of China (No.2022YFB3904700), Key Research and Development Program of in Shandong Province (2019JZZY020102), Key Research and Development Program of Jiangsu Province (No.BE2018084), Industrial Internet Innovation and Development Project in 2021 (TC210A02M, TC210804D), Opening Project of Beijing Key Laboratory of Mobile Computing and Pervasive Device.